\documentclass[runningheads]{llncs}
\usepackage{epsfig}
\usepackage{graphicx}
\usepackage{amsmath}
\usepackage{amssymb}
\usepackage{subcaption}
\usepackage{algorithm}
\usepackage{algorithmic}
\usepackage{comment}
\usepackage{textcomp}
\usepackage{amsmath,amsfonts,amssymb}
\usepackage{pifont}
\usepackage{booktabs}
\usepackage{bm}
\usepackage{multirow}
\usepackage{multicol}
\usepackage{enumitem}
\usepackage{xcolor}
\usepackage{url}
\usepackage{makecell}

\usepackage[T1]{fontenc}

\usepackage{hyphenat}
\newcommand{\repparams}{\theta}	
\newcommand{\taskparams}{W}

\newcommand{\task}{\mathcal{T}}

\newcommand{\Softmax}{\textit{Softmax}}

\newcommand{\tit}[1]{\smallbreak\noindent\textbf{#1}}
\newcommand{\tinytit}[1]{\noindent\textbf{#1}}
\usepackage{trimclip}
\newcommand{\shorttextleftarrow}{\clipbox*{0pt 0pt {.75\width} {\height}} \textleftarrow}

\usepackage{hyperref}

\begin{document}
\title{Generalising via Meta-Examples for Continual Learning in the Wild}
%
%

\author{Alessia Bertugli\inst{1} \and
Stefano Vincenzi\inst{2} \and
Simone Calderara\inst{2} \and Andrea Passerini\inst{1}}

\authorrunning{A. Bertugli et al.}

\institute{Università di Trento, Trento, Italy\\
\email{name.surname@unitn.it} \and
Università di Modena e Reggio Emilia, Modena, Italy \\
\email{name.surname@unimore.it}}

\maketitle              

\begin{abstract}
Future deep learning systems call for techniques that can deal with the evolving nature of temporal data and scarcity of annotations when new problems occur.
As a step towards this goal, we present FUSION (Few-shot UnSupervIsed cONtinual learning), a learning strategy that enables a neural network to learn quickly and continually on streams of unlabelled data and unbalanced tasks. The objective is to maximise the knowledge extracted from the unlabelled data stream (unsupervised), favor the forward transfer of previously learnt tasks and features (continual) and exploit as much as possible the supervised information when available (few-shot). The core of FUSION is MEML - Meta-Example Meta-Learning – that consolidates a meta-representation through the use of a self-attention mechanism during a single inner loop in the meta-optimisation stage. To further enhance the capability of MEML to generalise from few data, we extend it by creating various augmented surrogate tasks and by optimising over the hardest. 
An extensive experimental evaluation on public computer vision benchmarks shows that FUSION outperforms existing state-of-the-art solutions both in the few-shot and continual learning experimental settings. ~\footnote[1]{The code is available at \url{https://github.com/alessiabertugli/FUSION}}

\keywords{Continual Learning  \and Meta-Learning \and Representation Learning.}
\end{abstract}

\section{Introduction}
\label{sec:introduction}
Human-like learning has always been a challenge for deep learning algorithms. Neural networks work differently than the human brain, needing a large number of independent and identically distributed (iid) labelled data to face up the training process. Due to their weakness to directly deal with few, online, and unlabelled data, the majority of deep learning approaches are bounded to specific applications. Continual learning, meta-learning, and unsupervised learning try to overcome these limitations by proposing targeted solutions.
\begin{figure}[!t]

    \centering
    \includegraphics[width=7cm]{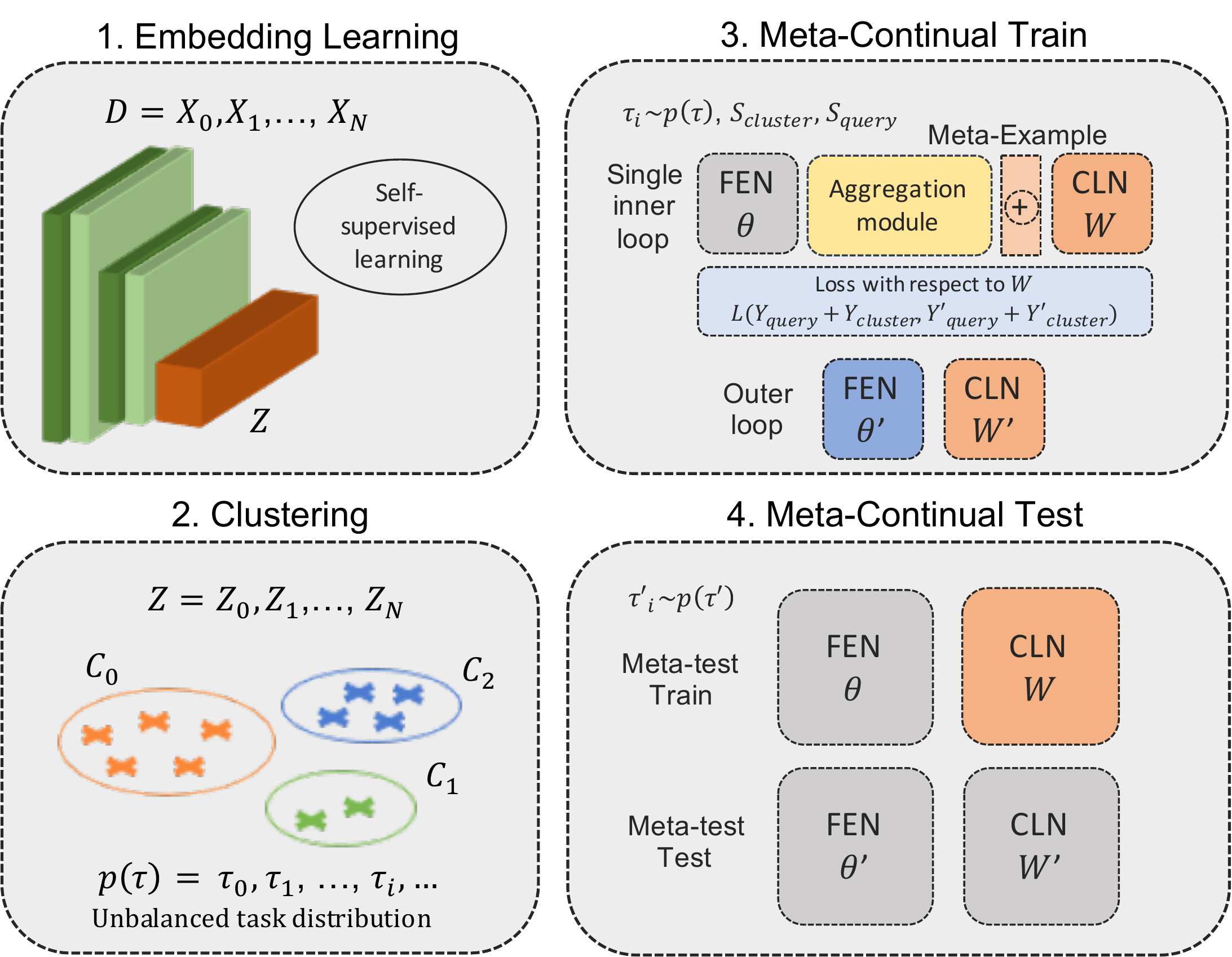}
    \caption{Overview of FUSION learning strategy. Further details in Section~\ref{sec:method}.}
    \label{fig:model}
\end{figure}
In particular, continual learning has been largely investigated in the last few years to solve the catastrophic forgetting problem that affects neural networks trained on incremental data. When data are available as a stream of tasks, neural networks tend to focus on the most recent, overwriting their past knowledge and consequently causing forgetting. Several methods~\cite{ewc,gem,a-gem,mer,si,iCaRL} have been proposed to solve this issue involving a memory buffer, network expansion, selective regularisation, and distillation. Some works take advantage of the meta-learning abilities of generalisation on different tasks and rapid learning on new ones to deal with continual learning problems, giving life to meta-continual learning~\cite{oml} and continual-meta learning~\cite{osaka}. Due to the complex nature of the problem, the proposed approaches generally involve supervised or reinforcement learning settings.
Moreover, the majority of continual learning solutions assume that data are perfectly balanced or equally distributed among classes. This problem is non-trivial for continual learning since specific solutions have to be found to preserve a balanced memory in presence of an imbalanced stream of data. 

In this paper, we introduce FUSION (standing for Few-shot UnSupervIsed cONtinual learning), a new learning strategy for unsupervised meta-continual learning that can learn from small datasets and does not require the underlying tasks to be balanced. 
As reported in Figure~\ref{fig:model}, FUSION is composed of four phases: embedding learning, clustering, meta-continual train and meta-continual test.
In the embedding learning phase, a neural network is trained to generate embeddings that facilitate the subsequent separation.
Then clustering is applied on these embeddings, and each cluster corresponds to a task (i.e., a class) for the following phase. As clustering is not constrained to produce balanced clusters, the resulting tasks are also unbalanced. For the meta-continual training phase, we introduce a novel meta-learning based algorithm that can effectively cope with unbalanced tasks. The algorithm, called MEML (for Meta-Example Meta-Learning), relies on a single inner loop update performed on an aggregated attentive representation, that we call meta-example. In so doing, MEML learns meta-representations that enrich the general features provided by large clusters with the variability given by small clusters, while existing approaches simply discard small clusters~\cite{cactus}. 
Finally, on meta-continual test, the learned representation is frozen and novel tasks are learned acting only on classifications layers.
We perform extensive experiments on two few-shot datasets, Omniglot~\cite{omniglot} and Mini-ImageNet~\cite{imagenet} and on two continual learning benchmarks, Sequential MNIST and Sequential CIFAR-10~\cite{cifar100} widely outperforming state-of-the-art methods.
\tit{Contributions.} We remark our contributions as follows:
\begin{itemize}[noitemsep,topsep=0pt]
    \item We propose FUSION, a novel strategy dealing with unbalanced tasks in an unsupervised meta-continual learning scenario;
    \item As part of FUSION, we introduce MEML, a new meta-learning based algorithm that can effectively cope with unbalanced tasks, and MEMLX, a variant of MEML exploiting an original augmentation technique to increase robustness, especially when dealing with undersized datasets;
    \item We test FUSION on an unsupervised meta-continual learning setting reaching superior performance compared to state-of-the-art approaches. Ablations studies empirically show that the imbalance in the task dimension does not negatively affect the performance, and no balancing technique is required;
    \item We additionally test MEML, our meta-continual learning method, in standard supervised continual learning, achieving better results with respect to specifically tailored solutions.
\end{itemize}

\section{Related Work}
\label{sec:related}
\subsection{Continual Learning}
\tinytit{Background}. Continual learning is one of the most challenging problems in deep learning research since neural networks are heavily affected by catastrophic forgetting when data are available as a stream of tasks. In more detail, neural networks tend to focus on the most recent tasks, overwriting their past knowledge and consequently causing forgetting.
As theoretically exposed in ~\cite{van2019three}, there are three main evaluation protocols for comparing methods' performance: Task-IL, Domain-IL and Class-IL.  Task-IL is the easiest scenario since task-identity is always provided, even at test-time; Domain-IL only needs to solve the current task, no task-identity is necessary; Class-IL instead intends to solve all tasks seen so far with no task identity given.
Much of the recent literature~\cite{oml,l2cl,der} is directed towards methods that do not require the detection of the task change. Our proposed approach follows this line of research, using a rehearsal technique to avoid forgetting without the need for task identity and targeted solutions to find them. 
Finally, continual learning methods can be divided into three main categories.

\tinytit{Architectural strategies}. They are based on specific architectures designed to mitigate catastrophic forgetting~\cite{pnn}. Progressive Neural Networks (PNN)~\cite{pnn} are based on parameter-freezing and network expansion, but suffers from a capacity problem because it implies adding a column to the neural network at each new task, so growing up the number of tasks training the neural network becomes more difficult due to exploding/vanishing gradient problems. 

\tinytit{Regularisation strategies}. They rely on putting regularisation terms into the loss function, promoting selective consolidation of important past weights~\cite{ewc,si}. Elastic Weights Consolidation (EWC)~\cite{ewc} uses a regularization term to control catastrophic forgetting by selectively constraining the model weights that are important for previous tasks through the computation of Fisher information matrix of weights importance. Synaptic Intelligence (SI)~\cite{si} can be considered as a variant of EWC, that computes weights importance online during SGD.

\tinytit{Rehearsal strategies}. Rehearsal strategies focus on retaining part of past information and periodically replaying it to the model to strengthen connections for memories, involving meta-learning~\cite{mer}, combination of rehearsal and regularisation strategies~\cite{gem,a-gem}, knowledge distillation~\cite{lwf} and generative replay~\cite{genrep}. Experience Replay~\cite{er} stores a limited amount of information of the past and then adds a further term to the loss that takes into account loss minimization on the buffer data, besides the current data. Meta-Experience Replay (MER)~\cite{mer} induces a meta-learning update in the process and integrate an experience replay buffer, updated with reservoir sampling, facilitating continual learning while maximizing transfer and minimizing interference. Gradient Episodic Memory (GEM)~\cite{gem} and its more efficient version A-GEM~\cite{a-gem} is a mix of regularization and rehearsal strategies.

\subsection{Meta-Learning}
\tinytit{Background}. Meta-learning, or learning to learn, aims to improve the neural networks ability to rapidly learn new tasks with few training samples. 
The tasks can comprise a variety of problems, such as classification, regression and reinforcement learning, but differently from Continual learning, training doesn't occur with incremental tasks and models are evaluated on new unseen tasks.
The majority of meta-learning approaches proposed in the literature are based on Model-Agnostic Meta-Learning (MAML)~\cite{maml}.

\tinytit{MAML}. By learning an effective parameter initialisation, with a double loop procedure, MAML limits the number of stochastic gradient descent steps required to learn new tasks, speeding up the adaptation process performed at meta-test time. The double loop procedure acts as follow: an inner loop that updates the parameters of the neural network to learn task-specific features and an outer loop generalizing to all tasks. 
The success of MAML is due to its model-agnostic nature and the limited number of parameters it requires. Nevertheless, it suffers from some limitations related to the amount of computational power it needs during the training phase. To solve this issue, the authors propose a further version, First Order MAML (FOMAML) that focus on removing the second derivative causing the need for large computational resources. 
ANIL~\cite{anil} investigates the success of MAML finding that it mostly depends on feature reuse rather than rapid learning. This way, the authors propose a slim version of MAML, removing almost all inner loops except for task-specific heads.

\tinytit{Unsupervised meta-learning}. Although MAML is suitable for many learning settings, few works investigate the unsupervised meta-learning problem. CACTUs~\cite{cactus} proposes a new unsupervised meta-learning method relying on clustering feature embeddings through the k-means algorithm and then builds tasks upon the predicted classes. The authors employ representation learning strategies to learn compliant embeddings during a pre-training phase. From these learned embeddings, a k-means algorithm clusters the features and assigns pseudo-labels to all samples. Finally, the tasks are built on these pseudo-labels.

\subsection{Meta-Learning for Continual Learning}
Meta-learning has been extensively merged with continual learning for different purposes. We highlight the existence of two strands of literature~\cite{osaka}: \emph{meta-continual learning}, that aims to incremental task learning, and \emph{continual-meta learning} that instead focuses on fast remembering.
To clarify the difference between these two branches, we adopt the standard notation that denotes $S$ as the support set and $Q$ as the query set. 
The sets are generated from the data distribution of the context (task) $C$ and respectively contain the samples employed in the inner and outer loops (e.g. in a classification scenario, both $S$ and $Q$ contain different samples of the same classes included in the current task). 
We define the meta-learning algorithm as $ML_{\phi}$ and the continual learning one with $CL_{\phi}$.  

\tinytit{Continual-meta learning.} Continual-meta learning mainly focuses on making meta-learning algorithms online, to rapidly remember meta-test tasks. 
In detail, it considers a sequence of tasks $S_{1:T}$, $Q_{1:T}$, where the inner loop computation is performed through $f_{\theta{t}} = ML_{\phi}(S_{t-1})$, while the learning of $\phi$ (outer loop) is obtained using gradient descent over the $l_{t} = \mathcal{L}(f_{\theta{t}}, S_{t})$. Since local stationarity is assumed, the model fails on its first prediction when the task switches. At the end of the sequence, $ML_{\phi}$ recomputes the inner loops over the previous supports and evaluate on the query set $Q_{1:T}$. 

\tinytit{Meta-continual learning.} More relevant to our work are meta-continual learning algorithms~\cite{oml,l2cl}, which use meta-learning rules to ``learn how not to forget". Resembling the notation proposed in~\cite{osaka}, given $K$ sequences sampled i.i.d. from a distribution of contexts $C$, $S_{i,1:T}, Q_{i,1:T} \sim X_{i,1:T} | C_{i,1:T}$, $CL_{\phi}$ is learned with $\nabla_{\phi} \sum_{t} \mathcal{L}(CL_{\phi}(S_{t}), Q_t)$ with $i < N < K$ and evaluated on the left out sets $\sum_{i=N}^K \mathcal{L}(CL_{\phi}(S_{t}), Q_t)$. 
In particular, OML~\cite{oml} and its variant ANML~\cite{l2cl} favour sparse representations by employing a trajectory-input update in the inner loop and a random-input update in the outer one. The algorithm jointly trains a representation learning network (RLN) and a prediction learning network (PLN) during the meta-training phase. Then, at meta-test time, the RLN layers are frozen and only the PLN is updated. ANML replaces the RLN network with a neuro-modulatory network that acts as a gating mechanism on the PLN activations following the idea of conditional computation.

\section{Few-Shot Unsupervised Continual Learning}
\label{sec:method}
Meta-continual learning~\cite{oml,l2cl} deals with the problem of allowing neural networks to learn from a stream of few, non i.i.d. examples and quickly adapt to new tasks. It can be considered as a few-shot learning problem, where tasks are incrementally seen, one class after the others. Formally, we define a distribution of training classification tasks $p(\task) = \task_{0}, \task_{1}, ..., \task_{i}, ...$. During meta-continual training, the neural network sees all samples belonging to $\task_{0}$ first, then all samples belonging to $\task_{1}$, and so on, without shuffling elements across tasks as in traditional deep learning settings. The network should be able to learn a general representation, capturing important features across tasks, without catastrophic forgetting, meaning to overfit on the last seen tasks. During the meta-test phase, a different distribution of unknown tasks $p(\task') = \task'_{0}, \task'_{1}, ..., \task'_{i}, ...$ is presented to the neural network again in an incremental way. The neural network, starting from the learned representation, should quickly learn to solve the novel tasks.
In this paper, differently from standard meta-continual learning, we focus on the case where no training labels are available and tasks have to be constructed in an unsupervised way, using pseudo-labels instead of the real labels in the meta-continual problem.
To investigate how neural networks learn when dealing with a real distribution and flow of unbalanced tasks, we propose FUSION, a novel learning strategy composed of four phases. 

\vspace{-0.2cm}
\subsection{Embedding Learning}
\vspace{-0.1cm}
Rather than requiring the task construction phase to directly work on high dimensional raw data, an embedding learning network, which is different from the one employed in the following phases, is used to determine an embedding that facilitates the subsequent task construction. Through an unsupervised training~\cite{deepcluster,acai}, the embedding learning network produces an embedding vector set $Z = Z_{0}, Z_{1},..., Z_{N}$, starting from the N data points in the training set $D = X_{0}, X_{1},..., X_{N}$ (see Figure~\ref{fig:model}.1). Embeddings can be learned in different ways, through generative models~\cite{acai} or self-supervised learning~\cite{deepcluster}. 
In Figure~\ref{fig:model}.1 an illustration of an unsupervised embedding learning based on self-supervised learning is shown. 
\vspace{-0.2cm}
\subsection{Clustering}
\vspace{-0.1cm}
As done in~\cite{cactus}, the task construction phase exploits the k-means algorithm over suitable embeddings obtained with the embedding learning phase previously described. This simple but effective method assigns the same pseudo-label to all data points belonging to the same cluster. This way, a distribution $p(\task) = \task_{0}, \task_{1}, ..., \task_{i}, ...$ of tasks is built from the generated clusters as reported in Figure~\ref{fig:model}.2.
Applying k-means over these embeddings leads to unbalanced clusters, which determine unbalanced tasks. This is in contrast with typical meta-learning and continual learning problems, where data are perfectly balanced. To recover a balanced setting, in~\cite{cactus}, the authors set a threshold on the cluster dimension, discarding extra samples and smaller clusters.
We believe that these approaches are sub-optimal as they alter the data distribution. In an unsupervised setting, where data points are grouped based on the similarity of their features, variability is an essential factor.
In a task imbalanced setting, the obtained meta-representation is influenced by both small and large clusters. 

\subsection{Meta-Continual Train}
\tit{Motivation.}
The adopted training protocol is related to the way data are provided at meta-test train time. In that phase, the model receives as input a stream of new unseen tasks, each with correlated samples; we do not assume access to other classes (opposed to the training phase) and only the current one is available. In this respect, since the network's finetuning occur with this stream of data, during training we reproduce a comparable scenario.
In particular, we need to design a training strategy that is sample efficient and directly optimize for a proper initial weights configuration. These suitable weights allow the network to work well on novel tasks after a few gradient steps using only a few samples. 
In the context of meta-learning, MAML relies on a two-loop training procedure performed on a batch of training tasks. The inner loop completes $N$ step of gradient updates on a portion of samples of the training tasks, while the outer loop exploits the remaining ones to optimize for a quickly adaptable representation (meta-objective).
Recent investigations on this algorithm explain that the real reason for MAML's success resides in feature reuse instead of rapid learning~\cite{anil}, proving that learning meaningful representations is a crucial factor.

\tit{Procedure.}
The created tasks are sampled one at a time $\task_{i} \sim p(\task)$ for the unsupervised meta-continual training phase as shown in Figure~\ref{fig:model}.3. The training process happens in a class-incremental way - where one task corresponds to one cluster - following a two-loop update procedure.
The inner loop involves samples belonging to the ongoing task, while the outer loop contains elements sampled from both the current and other random clusters.
In fact, during this stage, the network may suffer from the catastrophic forgetting effect on the learned representation if no technique is used to generalise or remember. To this end, the query set, used to update parameters in the outer loop, have to be designed to simulate an iid distribution, containing elements belonging to different tasks.
The unbalanced case takes two-third of the current cluster data for the inner loop and adds one-third to a fixed number of random samples for the outer loop. 
The balanced case - usually adopted with supervised data - instead takes the same number of samples among tasks for both the inner and the outer loop.
To deal with the meta-continual train in FUSION (Figure~\ref{fig:model}.3), we propose MEML, a meta-learning procedure based on the construction of a meta-example, a prototype of the task obtained through self-attention. The whole architecture is composed of a Feature Extraction Network (FEN), an aggregation module and a CLassification Network (CLN). The FEN is updated only in the outer loop (highlighted in blue in the figure), while frozen during the inner (grey). Both the aggregation module and the CLN are renewed in the inner and outer loop. 
\tit{MEML.}
We remove the need for several inner loops, maintaining a single inner loop update through a mechanism for aggregating examples based on self-attention. This way, we considerably reduce the training time and computational resources needed for training the model and increases global performance. The use of a meta-example instead of a trajectory of samples is particularly helpful in class-incremental continual learning to avoid catastrophic forgetting. In fact, instead of sequentially processing multiple examples of the same class and updating the parameters at each one (or at each batch), the network does it only once per class, reducing the forgetting effect.
At each time-step, a task $\task_{i} = (\mathcal S_{cluster}, \mathcal S_{query})$ is randomly sampled from the task distribution $p(\task)$. $\mathcal S_{cluster}$ contains elements of the same cluster as indicated in equation~\ref{eq:cluster}, where $Y_{cluster} = Y_{0} =... = Y_{k}$ is the cluster pseudo-label and $K$ is the number of data points in the cluster. Instead, $\mathcal S_{query}$ (equation~\ref{eq:query}) contains a variable number of elements belonging to the current cluster and a fixed number of elements randomly sampled from all other clusters, where $Q$ is the total number of elements in the query set.

\noindent\begin{minipage}{0.49\linewidth}
\begin{equation}
 \label{eq:cluster}
    \mathcal S_{cluster} = \{(X_{k}, Y_{k})\}_{k=0}^K, \;
\end{equation}
\end{minipage}
\begin{minipage}{0.49\linewidth}
\begin{equation}
 \label{eq:query}
    \mathcal S_{query} = \{(X_{q}, Y_{q})\}_{q=0}^Q.
\end{equation}
\end{minipage}
\vspace{0.2cm}


%
%
$\mathcal S_{cluster}$ is used for the inner loop update, while $\mathcal S_{query}$ is used to optimise the meta-objective during the outer loop. All the elements belonging to $\mathcal S_{cluster}$  are processed by the FEN, parameterised by $\theta$, computing the feature vectors $R_{0}, R_{1}, ..., R_{K}$ in parallel for all task elements (see equation~\ref{eq:fvec}). The obtained embeddings are refined in equation~\ref{eq:alpha} with an attention function, parameterised by $\rho$, that computes the attention coefficients $\vec{a}$ from the features vectors. Then, the final aggregated representation vector $R_{ME}$ (equation~\ref{eq:sum}), for \emph{meta-example} representation, captures the most salient features.

\begin{minipage}{0.30\linewidth}
    \begin{equation}
        \label{eq:fvec}
        R_{0:K} = f_\theta(X_{0:K}),
    \end{equation}
\end{minipage}
\begin{minipage}{0.36\linewidth}
    \begin{equation}
        \label{eq:alpha}
        \vec{a} = \Softmax[f_\rho(R_{0:K})],
    \end{equation}
\end{minipage}
\begin{minipage}{0.29\linewidth}
\begin{equation}
    \label{eq:sum}
    R_{ME} = \displaystyle \vec{a}^\intercal R_{0:K}.
\end{equation}
\end{minipage}

\vspace{0.2cm}
%
%
%
%
The single inner loop is performed on this meta-example, which adds up the weighted-features contribution of each element of the current cluster. Then, the cross-entropy loss $\mathcal{L}$ between the predicted label and the pseudo-label is computed and both the classification network parameters $W$ and the attention parameters $\rho$ are updated with a gradient descent step (as indicated in~\ref{eq:inner_loop}), where $\psi = \{\taskparams, \rho\}$ and $\alpha$ is the inner loop learning rate.
Finally, to update the whole network parameters $\phi =\{\repparams, \taskparams, \rho\}$, and to ensure generalisation across tasks, the outer loop loss is computed from $S_{query}$. The outer loop parameters are thus updated as shown in equation~\ref{eq:outer_loop} below, where $\beta$ is the outer loop learning rate.

\hspace{-0.6cm}
\begin{minipage}{0.51\linewidth}
    \begin{equation}
        \label{eq:inner_loop}
        \psi \leftarrow \psi-\alpha
            \nabla_{\psi} \mathcal{L}(f_{\psi}(R_{ME}), Y_{cluster}),
    \end{equation}
\end{minipage}
\begin{minipage}{0.49\linewidth}
    \begin{equation}
        \label{eq:outer_loop}
        \phi \leftarrow \phi - \alpha \nabla_\phi \mathcal{L} (f_{\phi}( X_{0:Q}), Y_{0:Q}).
    \end{equation}
\end{minipage}
\vspace{0.2cm}

%

Note that with the aggregation mechanism introduced by MEML, a single inner loop is made regardless of the number of examples in the cluster, thus eliminating the problem of unbalancing at the inner loop level.

\tit{MEMLX.}
Since the aim is to learn a representation that generalises to unseen classes, we introduce an original augmentation technique inspired by \cite{gong2020maxup}. The idea is to generate multiple sets of augmented input data and retain the set with maximal loss to be used as training data.
Minimising the average risk of this worst-case augmented data set enforces robustness and acts as a regularisation against random perturbations, leading to a boost in the generalisation capability. Starting from the previously defined $\mathcal S_{cluster}$ and $\mathcal S_{query}$ we generate $m$ sets of augmented data:
\begin{equation}
\label{eq:aug_maxup}
    \{\mathcal S_{cluster}^i, S_{query}^i\}_{i=1}^m \leftarrow A(\mathcal S_{cluster}), A(\mathcal S_{query}), \\
\end{equation}
where $A$ is an augmentation strategy that executes a combination of different data transformations for each $i \in m$.
Hence, for each of these newly generated sets of data we perform an evaluation forward pass through the network and compute the loss, retaining the $S_{cluster}^{i_c}$ and $\mathcal S_{query}^{i_q}$ sets giving the highest loss to be used as input to MEML for the training step: 
\begin{equation}
\begin{aligned}
\label{equ:maxup_loss}
i_c = \textrm{argmax}_{i\in 1,..m} \mathcal{L}(f(\mathcal S_{cluster}^i),Y_{cluster}), \\
i_q = \textrm{argmax}_{i\in 1,..m} \mathcal{L}(f(\mathcal S_{query}^i),Y_{0:Q}).
\end{aligned}
\end{equation}


%
Three different augmented batches are created starting from the input batch, each forwarded through the network producing logits. The Cross-Entropy losses between those latter and the targets are computed, keeping the augmented batch corresponding to the highest value.
In detail, we adopt the following augmentation: vertical flip, horizontal flip for batch 1; colour jitter (brightness, contrast, saturation, hue) for batch 2; random affine, random crop for batch 3.
\subsection{Meta-Continual Test}
At meta-continual test time, novel and unseen tasks $\task'_{i} \sim p(\task')$ from the test set are provided to the network,  as illustrated in Figure~\ref{fig:model}.4. Here $p(\task')$ represents the distribution of supervised test tasks and $\task'_{i}$ corresponds to a sampled test class. The representation learned during meta-train remains frozen, and only the prediction layers are fine-tuned. The test set is composed of novel tasks, that can be part of the same distribution (e.g. distinct classes within the same dataset) or even belong to a different distribution (e.g. training and testing performed on different datasets).

\vspace{-0.2cm}
\section{Experiments}
\label{sec:experiments}
\subsection{Few-Shot Unsupervised Continual Learning}

\tit{Datasets.}
We employ Omniglot and Mini-ImageNet, two datasets typically used for few-shot learning evaluation. The Omniglot dataset contains 1623 characters from 50 different alphabets with 20 greyscale image samples per class. We use the same splits as~\cite{cactus}, using 1100 characters for meta-training, 100 for meta-validation, and 423 for meta-testing.
The Mini-ImageNet dataset consists of 100 classes of realistic RGB images with 600 examples per class. We use 64 classes for meta-training, 16 for meta-validation and 20 for meta-test.

\tit{Architecture.} Following~\cite{oml}, we use for the FEN a six-layer CNN interleaved by ReLU activations with $256$ filters for Omniglot and $64$ for Mini-ImageNet. All convolutional layers have a $3\times3$ kernel (for Omniglot, the last one is a $1\times1$ kernel) and are followed by two linear layers constituting the CLN. The attention mechanism is implemented with two additional linear layers interleaved by a Tanh function, followed by a Softmax and a sum to compute attention coefficients and aggregate features. We use the same architecture for competitive methods. We do not apply the Softmax activation and the final aggregation but we keep the added linear layers, obtaining the same number of parameters. 
The choice in using two simple linear layers as attention mechanism is made specifically since the aim of the paper is to highlight how this kind of mechanism can enhance performance and significantly decrease both training time and memory usage. 
\tit{Training.}
For Omniglot, we train the model for $60000$ steps while for Mini-ImageNet for $200000$, with meta-batch size equals to $1$. The outer loop learning rate is set to $1e^{-4}$ while the inner loop learning rate is set to $0.1$ for Omniglot and $0.01$ for Mini-ImageNet, with Adam optimiser. As embedding learning networks, we employ Deep Cluster~\cite{deepcluster} for Mini-ImageNet and ACAI~\cite{acai} for Omniglot. Since Mini-ImageNet contains 600 examples per class, after clustering, we sample examples between 10 and 30, proportionally to the cluster dimension to keep the imbalance between tasks. We report the test accuracy on a different number of unseen classes, which induces increasingly complex problems as the number increase.
Following the protocol employed in~\cite{oml}, all results are obtained through the mean of $50$ runs for Omniglot and $5$ for Mini-ImageNet.
\begin{figure*}[t]
    \centering
    \begin{minipage}[b]{.56\linewidth}
        \resizebox{\linewidth}{!}{
             \setlength{\tabcolsep}{0.8mm}
             \renewcommand{\arraystretch}{0.9}
             \begin{tabular}{l|cccccc}
             \toprule\noalign{\smallskip}
             &\multicolumn{6}{c}{\textbf{Omniglot}} \\[0.1cm]
             Algorithm/Tasks   & 10 & 50 & 75 & 100 & 150 & 200 \\
             \noalign{\smallskip} \midrule
             Oracle OML~\cite{oml} & 88.4 & 74.0 & 69.8  & 57.4 & 51.6  & 47.9  \\ 
             Oracle ANML~\cite{l2cl} & 86.9 & 63.0 & 60.3 & 56.5 & 45.4 & 37.1 \\ 
             Oracle MEML (Ours) & \textbf{94.2} & \textbf{81.3} & \textbf{80.0} & \textbf{76.5} & \textbf{68.8} & \textbf{66.6} \\
             Oracle MEMLX (Ours) & \textbf{94.2} & 75.2 & 75.0 & 67.2 & 58.9 & 55.4 \\ 
             \midrule
             OML & 74.6 & 32.5 & 30.6 & 25.8 & 19.9 & 16.1 \\
             ANML & 72.2 & 46.5 & 43.7 & 37.9 & 26.5 & 20.8 \\ 
             MEML (Ours) & \textbf{89.0} & 48.9 & 46.6 & 37.0  & 29.3  & 25.9 \\
             MEMLX (Ours) & 82.8 & \textbf{50.6} & \textbf{49.8} & \textbf{42.0} & \textbf{34.9} & \textbf{31.0} \\
             \bottomrule
             \end{tabular}}
        \captionof{table}{Meta-test test accuracy on Omniglot.}
        \label{tab:omni}
    \end{minipage}\hfill \hspace{0.07cm}
    \begin{minipage}[b]{.40\linewidth}
        \resizebox{\linewidth}{!}{
            \centering
            \begin{tabular}{l|ccccc}
              \toprule\noalign{\smallskip}
             &\multicolumn{5}{c}{\textbf{Mini-ImageNet}} \\[0.1cm]
             Algorithm/Tasks   & 2 & 4 & 6 & 8 & 10 \\
             \noalign{\smallskip} \toprule
             Oracle OML~\cite{oml}  & 50.0 & 31.9 & 27.0  & 16.7  & 13.9  \\
             Oracle MEML (Ours) & 66.0 & 33.0 & 28.0 & 29.1 & 21.1 \\
             Oracle MEMLX (Ours) & \textbf{74.0} & \textbf{60.0} & \textbf{36.7} & \textbf{51.3} & \textbf{40.1} \\
             \midrule
             OML & 49.3 & 41.0  & 19.2  & 18.2 & 12.0 \\
             MEML (Ours) & 70.0  & \textbf{48.4} & 36.0  & 34.0 & 21.6  \\
             MEMLX (Ours) & \textbf{72.0} & 45.0 & \textbf{50.0} & \textbf{45.6} & \textbf{29.9} \\
             \bottomrule
             \end{tabular}}
      \captionof{table}{Meta-test test accuracy on Mini-ImageNet.}
     \label{tab:min}
     \end{minipage}
\end{figure*}
    
\begin{figure*}[t]
    \centering
    \begin{minipage}[b]{.52\linewidth}
        \raisebox{+1.8\height}{\resizebox{\linewidth}{!}{
             \setlength{\tabcolsep}{0.9mm}
             \renewcommand{\arraystretch}{1.1}
             \begin{tabular}{l|cc|cc}
             \toprule\noalign{\smallskip}
             &\multicolumn{2}{c|}{\textbf{Omniglot}} & \multicolumn{2}{c}{\textbf{Mini-ImageNet}}\\[0.1cm]
             Algorithm & Time & GPU & Time & GPU\\ \midrule
             OML~\cite{oml} & 1h 32m  & 2.239 GB  & 7h 44m & 3.111 GB \\
             MEML  & 47m & 0.743 GB & 3h 58m & 1.147 GB  \\ 
             MEMLX  & 1h 1m & 0.737 GB & 4h 52m & 1.149 GB\\ 
             \noalign{\smallskip} \bottomrule
             \end{tabular}}}
             \captionof{table}{Training time and GPU usage of MEML and MEMLX compared to OML on Omniglot and Mini-ImageNet.}
             \label{tab:time_memory}
    \end{minipage}\hfill \hspace{0.07cm}
    \begin{minipage}[b]{.4\linewidth}
        \resizebox{\linewidth}{!}{
            \centering
            \includegraphics[width=0.6\linewidth]{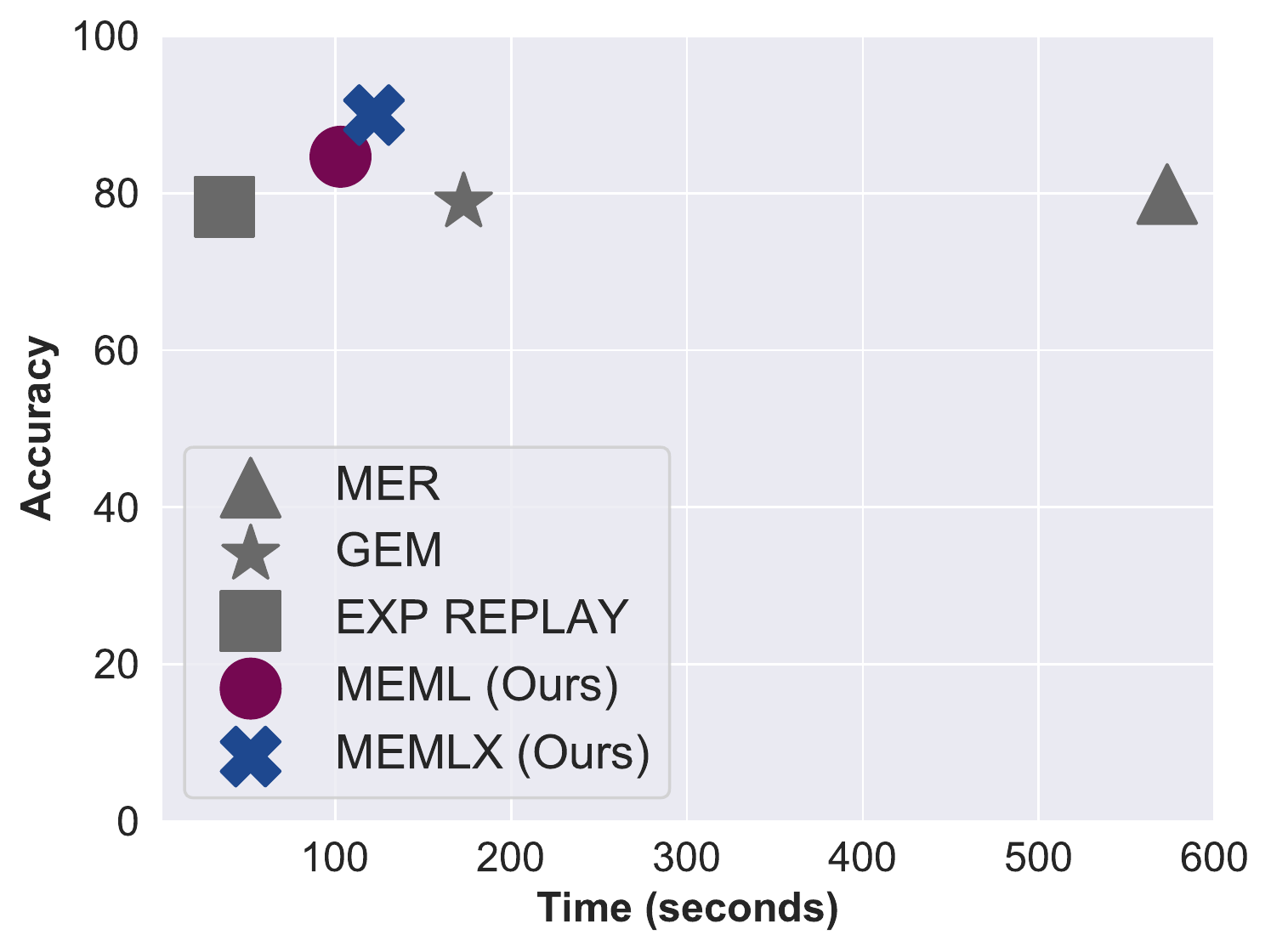}}
        \captionof{figure}{Training time comparison with respect to the accuracy between the most important state-of-the-art continual learning methods.}
        \label{fig:time_cl}
    \end{minipage}
\end{figure*}

\tit{Performance Analysis.}
In Table~\ref{tab:omni} and Table~\ref{tab:min}, we report results respectively on Omniglot and Mini-ImageNet, comparing our model with competing methods.
To see how the performance of MEML within our FUSION is far from those achievable with the real labels, we also report for all datasets the accuracy reached in a supervised case (\emph{oracles}).
We define Oracle OML~\cite{oml} and Oracle ANML~\cite{l2cl}~\footnote[2]{Our results on Oracle ANML are different from the ones presented in the original paper due to a different use of data. To make a fair comparison we use $10$ samples for the support set and $15$ for the query set for all models, while in the original ANML paper the authors use $20$ samples for the support set and $64$ for the query set. We do not test ANML on Mini-ImageNet due to the high computational resources needed.} as supervised competitors, and Oracle MEML the supervised version of our model.
MEML outperforms OML on Omniglot and Mini-ImageNet and ANML on Omniglot, suggesting that the meta-examples strategy is beneficial on both FUSION and fully supervised cases. MEMLX, the advanced version exploiting a specific augmentation technique is able to improve the MEML results in almost all experiments. In particular, MEMLX outperforms MEML on both Omniglot and Mini-ImageNet in FUSION and even in the fully supervised case on Mini-ImageNet.
The only experiment in which MEML outperforms MEMLX is on Omniglot, in the supervised case. In our opinion, the reason is to be found in the type of dataset. Omniglot is a dataset made up of 1100 classes and therefore characters are sometimes very similar to each other. Precisely for this reason, applying augmentation can lead the network to confuse augmented characters for a class with characters belonging to other classes. In the unsupervised case, the clusters are grouped by features, which should better separate the data from a visual point of view, thus favouring our augmentation technique.
\tit{Time and Computation Analysis.}
In Table~\ref{tab:time_memory}, we compare training time and computational resources usage between OML, MEML and MEMLX on Omniglot and Mini-ImageNet. We measure the time to complete all training steps and the computational resources in gigabytes occupied on the GPU. Both datasets confirm that our methods, adopting a single inner update, train considerably faster and uses approximately one-third of the GPU resources with respect to OML. MEMLX undergoes minimal slowdown, despite the use of our augmentation strategy. To a fair comparison, all tests are performed on the same hardware equipped with an NVIDIA Titan X GPU.

\vspace{-0.2cm}
\subsection{Supervised Continual Learning}

To further prove the effectiveness of our meta-example strategy, we put MEML and MEMLX in standard supervised continual learning and show its performance compared to state-of-the-art continual learning approaches.

\tit{Datasets.}
We experiment on Sequential MNIST and Sequential CIFAR-10.
In detail, the MNIST classification benchmark and the CIFAR-10 dataset~\cite{cifar100} are split into $5$ subsets of consecutive classes composed of $2$ classes each.

\tit{Architecture.}
For tests on Sequential MNIST, we employ as architecture a fully-connected network with two hidden layers, interleaving with ReLU activation as proposed in~\cite{gem},~\cite{mer}. For tests on CIFAR-10, we rely on ResNet18~\cite{iCaRL}. 

\tit{Training.}
We train all models in a class-incremental way (Class-IL), the hardest scenario among the three described in~\cite{van2019three}, which does not provide task identities. We train for $1$ epoch for Sequential MNIST and $50$ epochs for Sequential CIFAR-10. SGD optimiser is used for all methods for a fair comparison. A grid search of hyperparameters is performed on all models taking the best ones for each. 
For rehearsal-based strategies, we report results on buffer size $200$, $500$ and $5120$. The standard continual learning test protocol is used for all methods, where the accuracy is measured on test data composed of unseen samples of all training tasks at the end of the whole training process.
We adapt our meta-example strategy to a double class per task making a meta-example for each class corresponding to two inner loops. The query set used within FUSION mirror the memory buffer in continual learning. The memory buffer contains elements from previously seen tasks, while the query set samples elements from all training tasks. For MEMLX, we apply our augmentation technique on both current task data and buffer data.

\tit{Performance Analysis.}
In Table~\ref{tab:cl_tab} we show accuracy results on Sequential MNIST and Sequential CIFAR-10~\footnote[3]{Due to high training time we do not report MER results on Sequential CIFAR-10.} respectively.
MEML and MEMLX consistently overcome all state-of-the-art methods on both datasets. 
We denote that MEML is significantly different from MER, processing one sample at a time and making an inner loop on all samples. This greatly increases the training time, making this strategy ineffective for datasets such as CIFAR-10. On the contrary, MEML makes as many inner loops as there are classes per task and finally a single outer loop on both task data and buffer data. This way, MEML training time is comparable to the other rehearsal strategy, but with the generalisation benefit given from meta-learning.
To further confirm the beneficial role of the meta-learning procedure, we observe that EXP REPLAY, using only one loop, reaches lower performance.
In Table~\ref{tab:fbf_tab} we report results on additional continual learning metrics: forward transfer, backward transfer and forgetting. In particular, \textbf{forward transfer} (FWT) measure the capability of the model to improve on unseen tasks with respect to a random-initialized network. It is computed making the difference between the accuracy before the training on each task and the accuracy of a random-initialized network, averaged on all tasks. \textbf{Backward transfer}~\cite{gem} (BWT) is computed making the difference between the current accuracy and its best value for each task, making the assumption that the highest value of the accuracy on a task is the accuracy at the end of it. Finally, \textbf{forgetting} is similar to BTW, without the letter assumption. We compare the best performer algorithms on both Sequential MNIST and Sequential CIFAR. MEML and MEMLX outperform all the other methods on BWT and forgetting, while little lower performance are reached on FWT. Since results are consistent for all buffer dimensions, we report results on buffer $5120$. 
\begin{figure*}[t]
    \centering
    \begin{minipage}[b]{.5\linewidth}
        \resizebox{\linewidth}{!}{
            \setlength{\tabcolsep}{0.7mm}
            \renewcommand{\arraystretch}{1.}
            \begin{tabular}{l|cccc|cccc}
            \toprule\noalign{\smallskip}
            &\multicolumn{4}{c|}{\textbf{Sequential MNIST}} & \multicolumn{4}{c}{\textbf{Sequential CIFAR-10}}\\[0.1cm]
            Algorithm/Buffer & None & 200 & 500 & 5120 & None & 200 & 500 & 5120\\
            \noalign{\smallskip} \midrule
            LWF~\cite{lwf} & 19.62 & - & - & - & 19.60 & - & - & -\\
            EWC~\cite{ewc} & 20.07 & - & - & - & 19.52 & - & - & -\\
            SI~\cite{si} & 20.28 & - & - & - & 19.49 & - & - & -\\ 
            SAM~\cite{sam} & 62.63 & - & - & - & - & - & - & -\\
            iCARL~\cite{iCaRL} & -  &  - & - & - & - & 51.04 & 49.08 & 53.77 \\
            HAL~\cite{hal} & - & 79.80 & 86.80 & 88.68 & - & 32.72 & 46.24 & 66.26\\
            GEM~\cite{gem} & - &  78.85 & 85.86 & 95.72 & - & 28.91 & 23.81 & 25.26 \\
            EXP REPLAY~\cite{er}  & - & 78.23 & 88.67 & 94.52 & - & 47.88 & 59.01 & 83.65 \\
            MER~\cite{mer}  & - & 79.90 & 88.38 & 94.58 & - & - & - & -\\
            MEML (Ours) & - & 84.63 & 90.85 & \textbf{96.04} & - & \textbf{54.33} & \textbf{66.41} & 83.91 \\ 
            MEMLX (Ours) & - & \textbf{89.94} & \textbf{92.11}  & 94.88 & - & 51.98 & 63.25 & \textbf{83.95} \\ \bottomrule
            \end{tabular}}
        \captionof{table}{MEML and MEMLX compared to state-of-the-art continual learning methods on Sequential MNIST and Sequential CIFAR-10 in class-incremental learning.}
        \label{tab:cl_tab}
    \end{minipage}\hfill \hspace{0.07cm}
    \begin{minipage}[b]{.45\linewidth}
        \resizebox{\columnwidth}{!}{
            \setlength{\tabcolsep}{0.7mm}
            \renewcommand{\arraystretch}{1.}
            \begin{tabular}{l|ccc|ccc}
            \toprule\noalign{\smallskip}
            &\multicolumn{3}{c|}{\textbf{Sequential MNIST}} & \multicolumn{3}{c}{\textbf{Sequential CIFAR-10}}\\[0.1cm]
            Algorithm/Metric  & FWT & BWT & Forgetting & FWT & BWT & Forgetting \\
            \noalign{\smallskip} \midrule
            HAL~\cite{hal} & -10.06 & -6.55 & 6.55 & -10.34 & -27.19 & 27.19 \\
            GEM~\cite{gem} & \textbf{-9.51} & -4.14 & 4.30 & -9.18 & -75.27 & 75.27 \\
            EXP REPLAY~\cite{er} & -10.97 & -6.07 & 6.08 & \textbf{-8.45} & -13.99 & 13.99 \\
            MER~\cite{mer} & -10.50 & -3.22 & 3.22 & - & - & - \\
            MEML (Ours) & -9.74 & -3.12 & 3.12 & -12.68 & \textbf{-10.97} & \textbf{10.97} \\ 
            MEMLX (Ours) & -9.74 & \textbf{-1.72} & \textbf{1.92} & -12.74 & -12.42 & 12.42 \\ \bottomrule
            \end{tabular}}
         \captionof{table}{Forward transfer, backward transfer and forgetting comparison on Sequential MNIST and Sequential CIFAR-10 in class-incremental learning.}
         \label{tab:fbf_tab}
    \end{minipage}
\end{figure*}



     

\tit{Time Analysis.}
We make a training time analysis (see Figure~\ref{fig:time_cl}) 
between the most relevant state-of-the-art continual learning strategy on Sequential MNIST. We measure the training time in seconds since the last task. We find that MEML and MEMLX are slower only compared to EXP REPLAY due to the meta-learning strategy, but they are faster with respect to both GEM and MER, reaching higher accuracy.

\begin{figure*}[!t]
\begin{minipage}[b]{4cm}
\centering
\captionsetup{width=1.2\linewidth}
\includegraphics[height=4.cm]{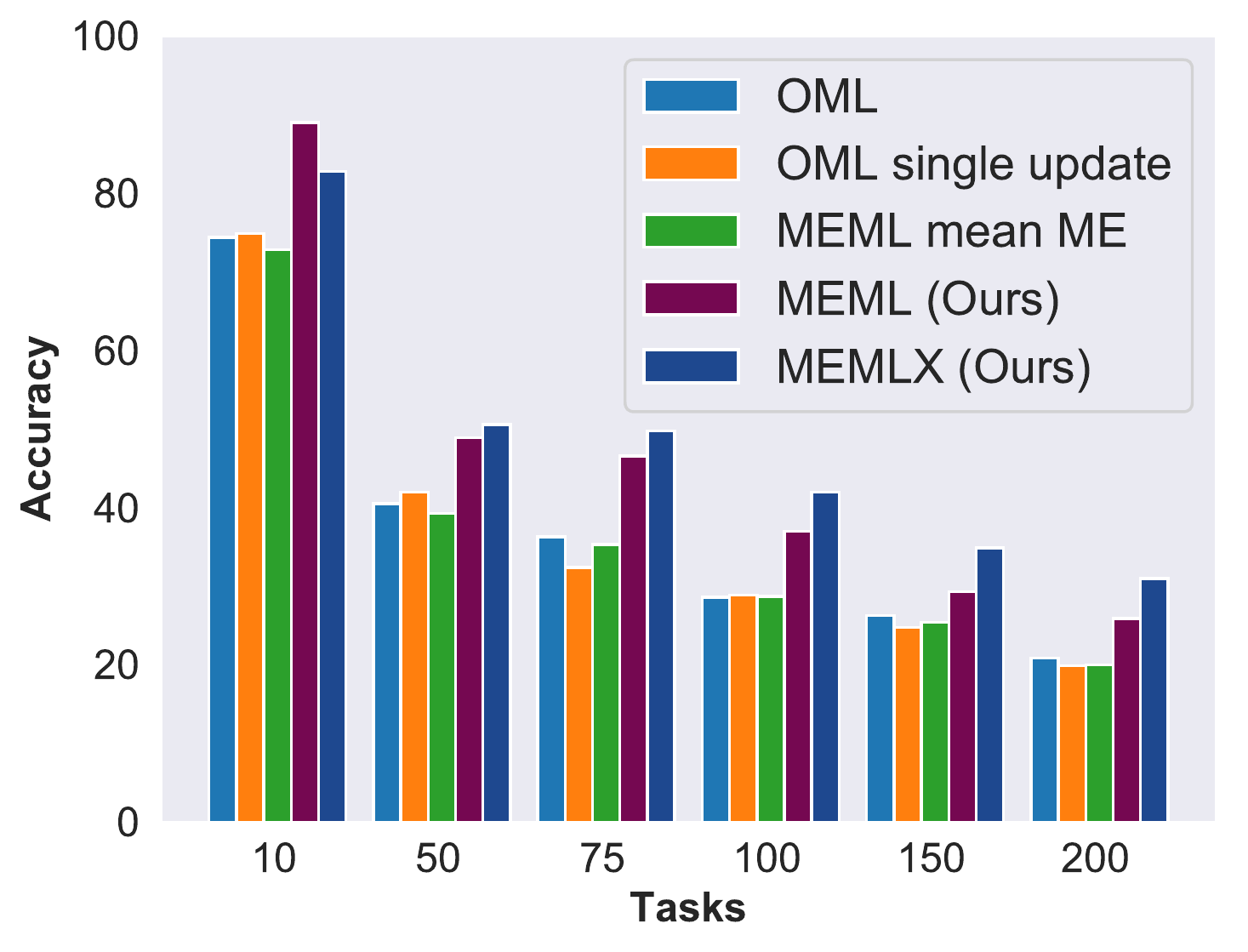} 
\caption{The capability of meta-example on Omniglot.}
\label{fig:me_abl}
\end{minipage}
\ \hspace{0.5mm} \hspace{1.5cm} \
\begin{minipage}[b]{4cm}
\centering
\captionsetup{width=1.2\linewidth}
\includegraphics[height=4.cm]{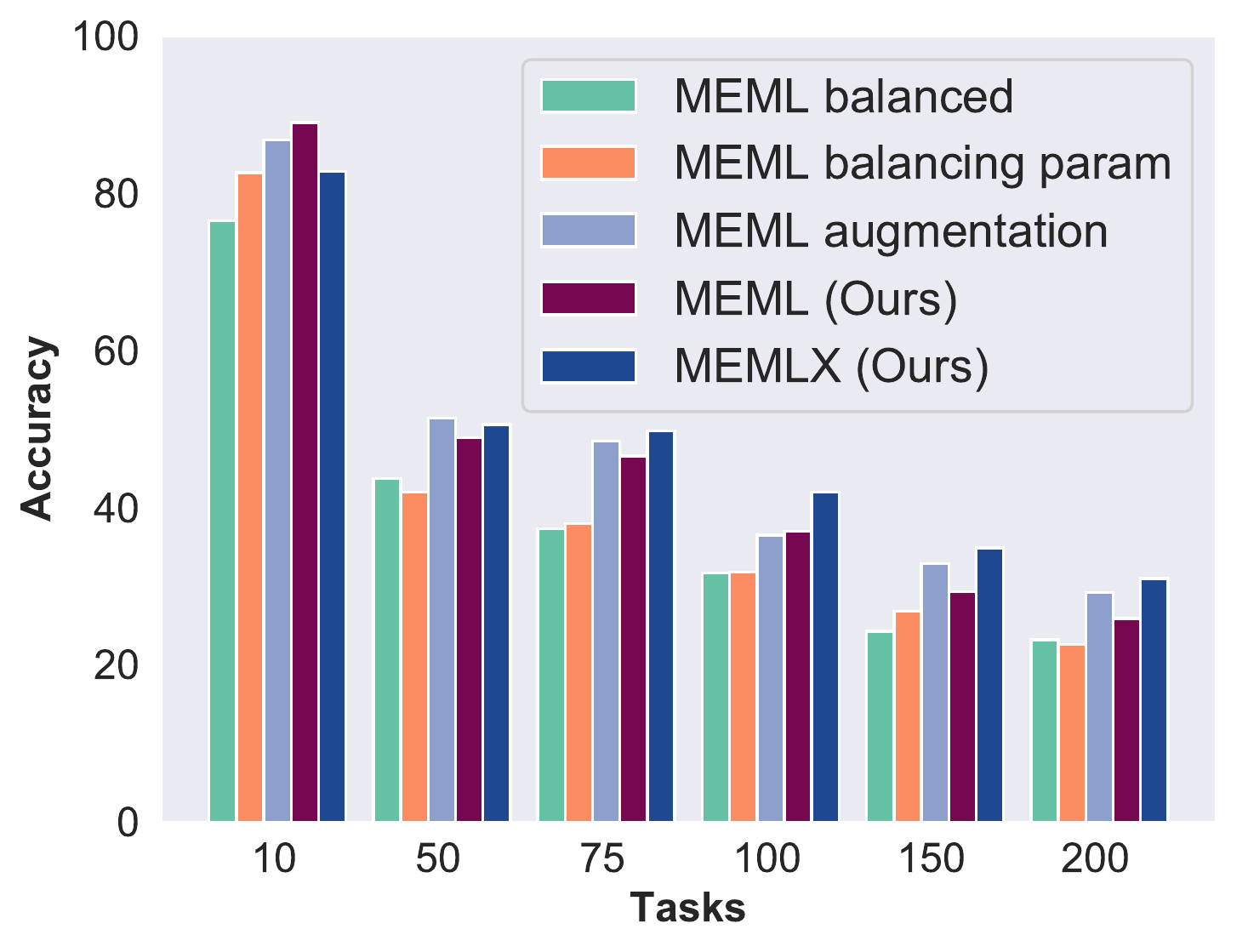}
\caption{Unbalanced \textit{vs.} balanced settings on Omniglot.}
\label{fig:unbal_abl}
\end{minipage} 
\end{figure*}

\tit{Meta-Example Single Update \textit{vs.} Multiple Updates.}
To prove the effectiveness of our method - MEML - based on meta-examples, we compare it with: OML~\cite{oml} - performing multiple updates, one for each element of the cluster; OML with a single update - adopting a single update over a randomly sampled data point from each task; MEML with mean ME - a version exploiting the mean between the feature vector computed by the FEN.
In Figure~\ref{fig:me_abl}, we show that MEML and MEMLX consistently outperform all the other baselines on Omniglot. 
OML with a single update gives analogous performance to the multiple updates one, confirming the idea that the strength of generalisation relies on the feature reuse. Also, the MEML with mean ME has performance comparable with the multiple and single update ones, proving the effectiveness of our aggregation mechanism to determine a suitable and general embedding vector for the CLN.

\tit{Balanced \textit{vs.} Unbalanced Tasks.}
To justify the use of unbalanced tasks and show that allowing unbalanced clusters is more beneficial than enforcing fewer balanced ones, we present some comparisons in Figure~\ref{fig:unbal_abl}.
First of all, we introduce a baseline in which the number of clusters is set to the true number of classes, removing from the task distribution the ones containing less than $N$ elements and sampling $N$ elements from the bigger ones. We thus obtain a perfectly balanced training set at the cost of less variety within the clusters; however, this leads to poor performance as small clusters are never represented. 
To verify if maintaining variety and balancing data can lead to better performance, we try two balancing strategies: augmentation, at data-level, and balancing parameter, at model-level. For the first one, we keep all clusters, sampling $N$ elements from the bigger and using data augmentation for the smaller to reach $N$ elements. 
At model-level, we multiply the loss term by a balancing parameter to weigh the update for each task based on cluster length.
These tests result in lower performance with respect to MEML and MEMLX, suggesting that the only thing that matters is cluster variety and unbalancing does not negatively affect the training. 

\vspace{-0.2cm}
\section{Conclusion and Future Work}
\label{sec:discussion}
We tackle a novel problem concerning few-shot unsupervised continual learning, proposing an effective learning strategy based on the construction of unbalanced tasks and meta-examples.
With an unconstrained clustering approach, we find that no balancing technique is necessary for an unsupervised scenario that needs to generalise to new tasks.
Our model, exploiting a single inner update through meta-examples, increase performance as the most relevant features are selected. In addition, an original augmentation technique is applied to reinforce its strength.
We show that MEML and MEMLX not only outperform the other baselines within FUSION but also exceed state-of-the-art approaches in class-incremental continual learning. 
Interesting future research is to investigate a more effective rehearsal strategy that further improves performance even when facing Out-of-Distribution data and domain shift.

%
%
\bibliographystyle{splncs04}
\bibliography{egbib}

\end{document}